\documentclass[10pt,twocolumn,letterpaper]{article}

\usepackage{iccv}              

\definecolor{iccvblue}{rgb}{0.21,0.49,0.74}
\usepackage[pagebackref,breaklinks,colorlinks,allcolors=iccvblue]{hyperref}

\usepackage{rotating}
\usepackage{multirow}

\usepackage[linesnumbered,ruled,vlined]{algorithm2e}
\usepackage{algorithmic} %

\def\paperID{2688} 
\def\confName{ICCV}
\def\confYear{2025}

\title{DuCos: Duality Constrained Depth Super-Resolution via Foundation Model}

\author{Zhiqiang Yan$^{1}$ \quad Zhengxue Wang$^{2}$ \quad Haoye Dong$^{1}$ \quad Jun Li$^{2}$ \quad Jian Yang$^{2}$ \quad Gim Hee Lee$^{1}$\\
$^1$National University of Singapore\quad
$^2$Nanjing University of Science and Technology\\
{\tt\small \{yanzq, gimhee.lee\}@nus.edu.sg}
}

\begin{document}
\maketitle
\begin{abstract}
We introduce DuCos, a novel depth super-resolution framework grounded in Lagrangian duality theory, offering a flexible integration of multiple constraints and reconstruction objectives to enhance accuracy and robustness. Our DuCos is the first to significantly improve generalization across diverse scenarios with foundation models as prompts. The prompt design consists of two key components: Correlative Fusion (CF) and Gradient Regulation (GR). CF facilitates precise geometric alignment and effective fusion between prompt and depth features, while GR refines depth predictions by enforcing consistency with sharp-edged depth maps derived from foundation models. Crucially, these prompts are seamlessly embedded into the Lagrangian constraint term, forming a synergistic and principled framework. Extensive experiments demonstrate that DuCos outperforms existing state-of-the-art methods, achieving superior accuracy, robustness, and generalization. The code is available at \url{https://github.com/yanzq95/DuCos}. 

\end{abstract}    
\section{Introduction}
\label{sec:introduction}

Depth super-resolution (DSR) \cite{zhong2023deep} is a fundamental task in computer vision that aims to restore high-resolution (HR) depth data from low-resolution (LR) depth inputs. It plays a crucial role in various downstream applications such as 3D reconstruction \cite{he2021towards,yan2022learning,de2022learning,yan2022rignet,liang2025distilling,zhou2024bring,zhou2024adverse}, augmented reality \cite{su2019pixel,wang2023rgb,yan2023distortion,yuan2023structure}, and robotic sensing \cite{sun2021learning,zhao2022discrete,metzger2023guided,yan2025completion}. These applications are largely reliant on clear and HR depth predictions. However, depth data typically has much lower resolution than color images \footnote{For example, the RGB-D system of Huawei P30 Pro \cite{yan2025rignet++,yan2024tri,he2021towards} captures $3648 \times 2736$ color images while the depth maps are only $240 \times 180$.} due to the differing advancements of color cameras and depth sensors. It is therefore essential to enhance the resolution of low-quality depth data. 

\begin{figure}[t]
 \centering
 \includegraphics[width=0.96\columnwidth]{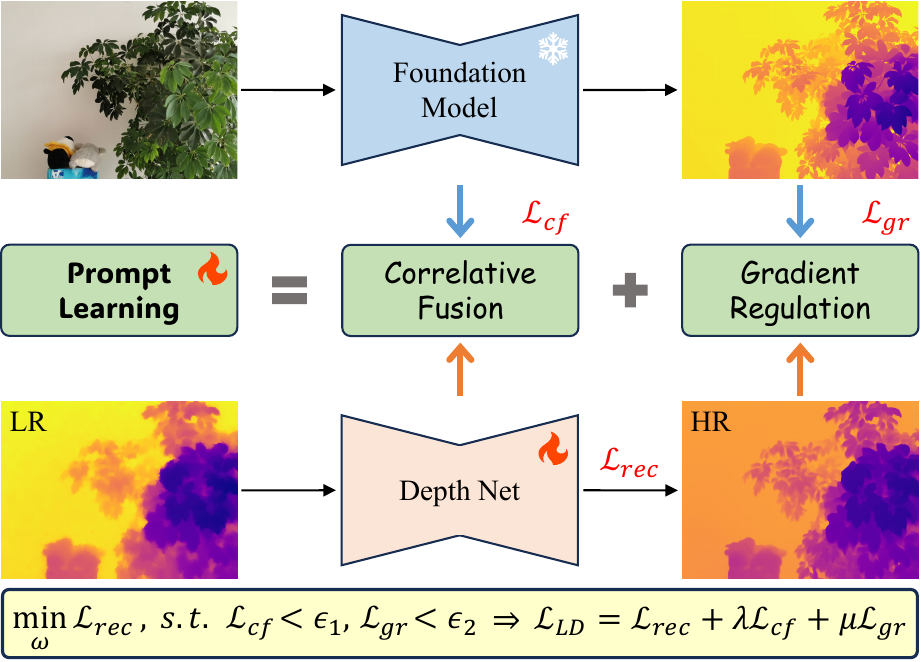}\\
 \vspace{-5pt}
 \caption{Overview of our \textbf{DuCos}: A highly synergistic system that integrates prompt learning with a Lagrangian duality (LD) based optimization algorithm. It combines the reconstruction loss with geometric consistency and edge preservation constraints to capture both global content and fine-grained local details, and uses foundation models to enhance generalization. 
 }\label{fig_overview}
  \vspace{-10pt}
\end{figure}

Numerous advanced approaches have been developed, often using a reconstruction loss to minimize the discrepancy between depth predictions and ground truth annotations. However, this loss focuses predominantly on global error minimization that relies heavily on pixel-level differences \cite{sun2021learning,zhao2022discrete,wang2024sgnet}. As a result, it often fails to capture fine-grained local geometric details, leading to blurred or oversmoothed depth recovery \cite{pan2019spatially,kim2021deformable,yan2022learning}. 
Moreover, the predictions become increasingly inaccurate and unreliable with disturbances such as noise in the input data. In addition, existing DSR solutions often focus on specialized network designs tailored to specific training data distributions. Despite their effectiveness in certain scenarios, these methods may face challenges in generalization across diverse resolutions and conditions. 

We propose \textbf{DuCos}, a novel paradigm of \textbf{du}ality \textbf{co}nstrained D\textbf{S}R via foundation model to address the above-mentioned issues. As illustrated in Fig.~\ref{fig_overview}, DuCos establishes a theoretical framework for accurate and robust DSR while integrating prompt learning to enhance generalization. Specifically, our DuCos first reformulates DSR as a constrained optimization problem with the Lagrangian duality (LD) theory \cite{boyd2004convex, palomar2006tutorial}. The total LD loss comprises a reconstruction term and a constraint term. The reconstruction term minimizes global errors, while the constraint term enforces geometric consistency and edge preservation. Guided by theoretical principles, this design allows the model to balance global context and local details, enhancing both accuracy and robustness. Furthermore, our DuCos is a pioneer in leveraging depth foundation models \cite{yang2024depth,hu2024metric3d,piccinelli2024unidepth} as prompts to boost generalization. Concretely, these prompts consist of correlation fusion (CF) and gradient regulation (GR). CF quantifies the similarity between RGB-D features using the Pearson Correlation Coefficient and subsequently conducts recurrent fusion. Such explicit data statistics facilitate intuitive and effective modeling. 
In addition, while depth predictions from depth foundation models often exhibit significant errors, they retain remarkably sharp edges. To take advantage of this property, GR computes their gradients to highlight edges and subsequently minimizes the normalized gradient errors, leading to improved accuracy. 

We further introduce an alignment loss in CF to enhance the geometric consistency between RGB-D features, and an edge-aware loss is designed in GR to promote edge preservation. These two loss functions are seamlessly integrated into the Lagrangian duality framework as the Lagrangian constraint term, resulting in a highly synergistic system capable of effectively capturing the sparsity and structural characteristics of depth maps. 

In summary, our contributions are as follows:
\begin{itemize}
\item 
We present a novel DSR paradigm based on Lagrangian duality theory, reformulating it as a constrained optimization problem to ensure a precise solution with rigorous theoretical foundations.
\item 
To our best knowledge, we are first to leverage foundation models as DSR prompts and seamlessly integrate them into the Lagrangian duality framework through CF and GR along with their respective constraints.
\item 
Extensive experiments demonstrate the superiority of our approach over state-of-the-art methods in terms of accuracy, robustness, and generalization on five datasets. 
\end{itemize}

\section{Related Work}
\label{sec:related_work}

\noindent
\textbf{Depth Super-Resolution.} 
Existing methods can be broadly categorized into synthetic DSR and real-world DSR, depending on the nature of the data used. For synthetic DSR, the degradation pattern is well defined~\cite{pan2019spatially,tang2021bridgenet,zhao2022discrete,metzger2023guided,wang2024scene,chen2024intrinsic,kang2025c2pd,chen2024intrinsic,wang2023learning} since LR depth inputs are typically generated from depth annotations using bicubic interpolation. 
For example, ATGVNet~\cite{riegler2016atgv} combines convolutional neural networks with total variation to produce HR depth maps. 
MSGNet~\cite{hui2016depth} introduces a multi-scale guidance network to enhance depth boundaries. PMBANet~\cite{ye2020pmbanet} presents a progressive multi-branch fusion network aimed at restoring high-resolution depth while maintaining sharp boundaries. Similarly, SSDNet~\cite{zhao2022discrete,zhao2023spherical} incorporates discrete cosine transform and spherical contrast refinement to address issues related to fine details. 
Recently, several guided image filtering approaches~\cite{li2019joint,kim2021deformable,zhong2023deep,wang2024sgnet} have been proposed to improve the transfer of guidance information into target depth maps. For example, DKN~\cite{kim2021deformable} utilizes deformable kernel networks to model sparse and spatially variant filter kernels, thereby improving the flexibility of filtering. Furthermore, advanced transformer~\cite{zhao2023spherical} and diffusion~\cite{metzger2023guided} techniques are also incorporated to further advance the DSR task. 
With the increasing technical and practical demands, real-world DSR with unknown degradation patterns emerges. For the first time, FDSR~\cite{he2021towards} builds a real-world benchmark dataset and a new blind DSR baseline. Subsequently, SFGNet~\cite{yuan2023structure} introduces a structure flow-guided model that learns cross-modal flows to effectively propagate the structural information of color images. Most recently, DORNet~\cite{wang2025dornet} utilizes self-supervised degradation learning for the first time to model the degradation patterns of real-world data, contributing to significant performance improvements. Unlike these methods that are optimized using a fixed reconstruction loss function, our objective is to develop a highly precise and robust model with stronger theoretical guarantees. 

\vspace{1mm}
\noindent
\textbf{Depth Prompt Learning.} 
A variety of depth estimation foundation models~\cite{yang2024depth,yang2025depth,yin2023metric3d,hu2024metric3d,piccinelli2024unidepth,ke2024repurposing} from a single image have been proposed. These models can provide strong priors for various depth perception tasks, including depth matching~\cite{jiang2025defom,wen2025foundationstereo,zhou2023unsupervised} and depth completion~\cite{park2024depth,park2024simple,viola2024marigold,liu2024depthlab,marsal2024foundation,lin2024prompting}. 
DEFOM-Stereo integrates Depth Anything v2~\cite{yang2025depth} into RAFT-Stereo~\cite{lipson2021raft} to enhance stereo matching capabilities. 
FoundationStereo~\cite{wen2025foundationstereo} enables robust and accurate zero-shot stereo depth estimation, bridging the gap between simulation and reality. To complete sparse depth maps, DepthPrompting~\cite{park2024depth} and UniDC~\cite{park2024simple} design a prompt mechanism that aggregates RGB priors from depth foundation models with depth features from depth-private branches. Marigold-DC~\cite{viola2024marigold} injects 
sparse depth observations as test-time guidance into a pretrained latent diffusion model for monocular depth estimation~\cite{ke2024repurposing}. Furthermore, DepthLab~\cite{liu2024depthlab} introduces image-conditioned depth inpainting based on diffusion priors, thereby facilitating various downstream applications. DepthRescaling~\cite{marsal2024foundation} proposes to rescale Depth Anything predictions using 3D points provided
by low-cost depth sensors or techniques. What's more, PromptDA~\cite{lin2024prompting} achieves 4K resolution depth estimation from low-quality depth data through concise prompt fusion and scaling designs. All of these depth prompt learning methods provide valuable insights into harnessing the powerful prior knowledge of depth foundation models to benefit image-guided depth perception tasks. 

 \begin{figure*}[t]
  \centering
  \includegraphics[width=1.8\columnwidth]{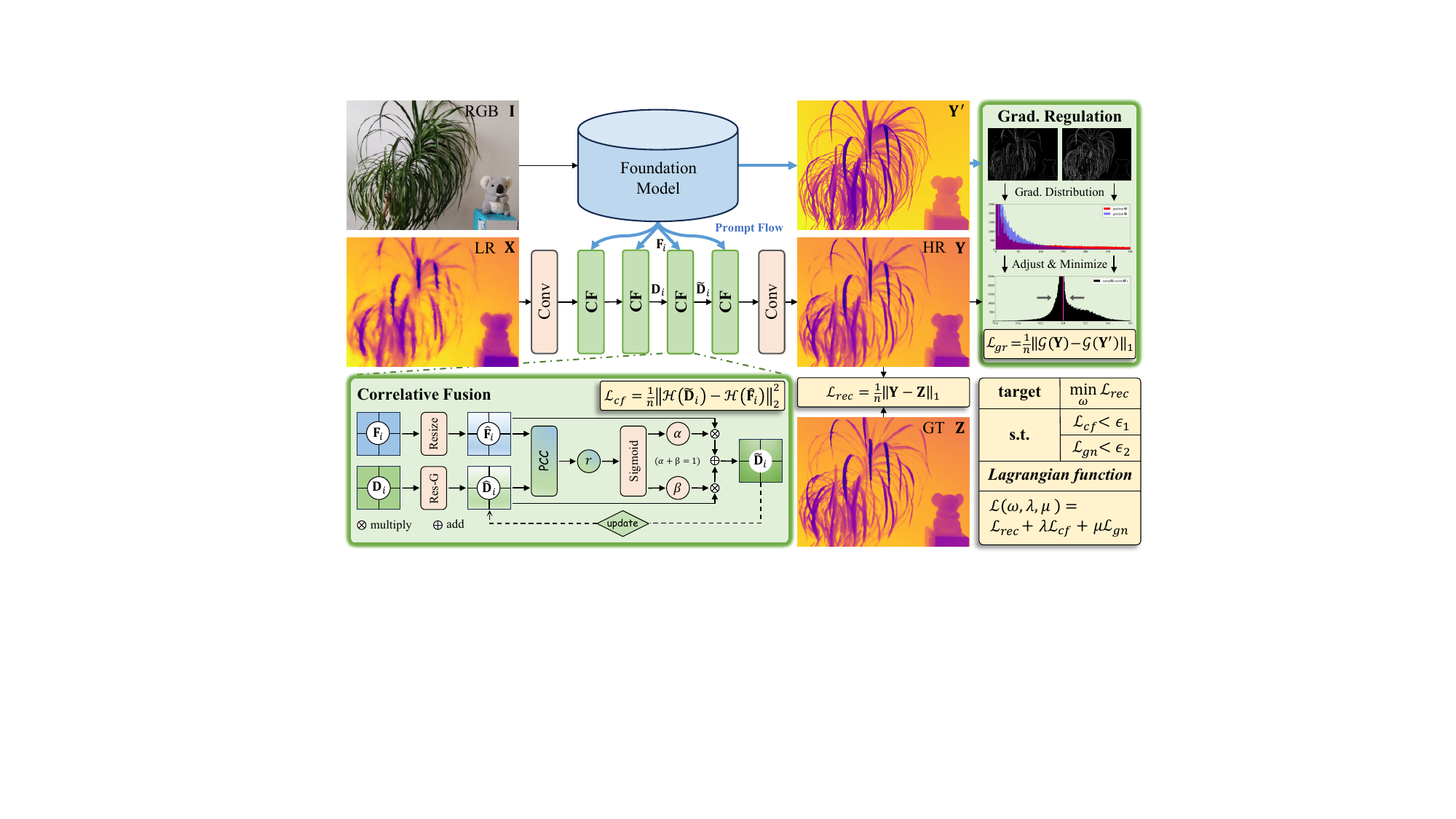}\\
   \vspace{-3pt}
  \caption{Detailed pipeline of DuCos. The color image $\mathbf{I}$ is first processed by a depth foundation model to generate prompt flows, including the intermediate features $\mathbf{F}_i$ and sharp-edged relative depth output $\mathbf{Y}'$. $\mathbf{F}_i$ is then used to guide depth recovery with Correlative Fusion (CF), and $\mathbf{Y}'$ enhances edges through Gradient Regulation (GR). Moreover, CF and GR are both formulated as constraints on reconstruction loss within the Lagrangian duality framework. Res-G: residual group~\cite{zhang2018image}, PCC: Pearson Correlation Coefficient.
  }\label{fig_pipeline}
   \vspace{-3pt}
\end{figure*}

\section{Our Method: DuCos}
\noindent \textbf{Overview.} Existing DSR methods focus mainly on specialized network designs. In contrast, our approach seeks to establish a general baseline by harnessing the strong priors of depth foundation models, guided by theoretical principles. 
%
%
Fig.~\ref{fig_pipeline} shows the pipeline of our DuCos with two key components: prompt learning and Lagrangian duality optimization. 
For prompt learning, the prompt flow $\mathbf{F}_i$ and $\mathbf{Y}'$ is derived from a depth foundation model given a color image $\mathbf{I}$. To effectively integrate these prompts, we first design correlative fusion (CF), which iteratively fuses the RGB prompt feature $\mathbf{F}_i$ and the depth feature $\mathbf{D}_i$ by computing their correlation. At the same time, an alignment loss is introduced to reinforce geometric consistency in RGB-D features. Next, we incorporate gradient regulation (GR) with an edge-aware loss to minimize the gradient discrepancies between the HR depth $\mathbf{Y}$ and the sharp-edged depth $\mathbf{Y}'$. 
For Lagrangian duality optimization, the auxiliary alignment and edge-aware losses serve as constraint terms, seamlessly integrated with the primary reconstruction loss using Lagrangian duality theory \cite{boyd2004convex,palomar2006tutorial}.

We give the details of our DuCos in the following sections. Sec.~\ref{sec_prompt_learning} elaborates on prompt learning and the corresponding restricted conditions. Sec.~\ref{sec_lagrangian_duality} describes the formulation of DSR as a constrained optimization problem under these restricted conditions, leading to its dual problem which is ultimately solved using Lagrangian duality theory.

\subsection{Prompt Learning}
\label{sec_prompt_learning}

\noindent \textbf{Correlative Fusion.}
As illustrated in Fig.~\ref{fig_pipeline}, the image branch processes the RGB input $\mathbf{I} \in \mathbb{R}^{3 \times H \times W}$ through a depth foundation model. A notable example is Depth Anything v2 \cite{yang2025depth}, which extracts the prompt flow $\mathbf{F}_i \in \mathbb{R}^{C \times H/p \times W/p}$ across four stages. We follow~\cite{yang2025depth} to set the patch size $p$ as 14. In the depth branch, the initial low-quality depth map undergoes upsampling by bicubic interpolation~\cite{he2021towards,zhao2022discrete,wang2024sgnet}, producing the depth input $\mathbf{X} \in \mathbb{R}^{1 \times H \times W}$. Subsequently, this input is encoded using a $3 \times 3$ convolutional head $\mathcal{F}_{{\tau}_1}(\cdot)$ to generate the feature representation $\mathbf{D}_1 \in \mathbb{R}^{C \times H \times W}$, \ie: 
\begin{subequations}
\begin{align}
    &\mathbf{F}_{i}=\mathcal{F}_{\phi} (\mathbf{I}), \quad \text{for} \quad 1 \leq i \leq 4; \\
    &\mathbf{D}_{1}=\mathcal{F}_{{\tau}_1}(\mathbf{X}),
\end{align}\label{eq_begin}
\end{subequations}
where $\mathcal{F}_{\phi}(\cdot)$ denotes the depth foundation model.

In the $i$-th stage, we introduce the correlative fusion module to effectively integrate the RGB-D features $\mathbf{F}_i$ and $\mathbf{D}_i$. Specifically, the prompt $\mathbf{F}_i$ is resized by deconvolution and interpolation to maintain RGB-D scale consistency, resulting in $\hat{\mathbf{F}}_i \in \mathbb{R}^{C \times H \times W}$. A residual group~\cite{zhang2018image} is then applied for further feature extraction, yielding $\hat{\mathbf{D}}_i \in \mathbb{R}^{C \times H \times W}$. This process can be denoted as:
\begin{subequations}
\begin{align}
    &\hat{\mathbf{F}}_{i}=\mathcal{F}_{\delta} (\mathbf{F}_i), \quad \text{for} \quad 1 \leq i \leq 4; \label{eq_cf_input_rgb}\\
    &\hat{\mathbf{D}}_{i}=\mathcal{F}_{\theta}(\mathbf{D}_i),\label{eq_cf_input_dep}
\end{align}
\end{subequations}
where $\mathcal{F}_{\delta}(\cdot)$ refers to the combined deconvolution and interpolation and $\mathcal{F}_{\theta}(\cdot)$ represents the residual group~\cite{zhang2018image}. Next, we utilize the statistical measure of Pearson Correlation Coefficient to explicitly quantify the relevance between the two features. Formally, the relevance is given by: 
\begin{equation}
    r = \frac{\sum (\hat{\mathbf{F}}_{i}^{j} - \bar{\mathbf{F}}_{i})(\hat{\mathbf{D}}_{i}^{j} - \bar{\mathbf{D}}_{i})}{\sqrt{\sum (\hat{\mathbf{F}}_{i}^{j} - \bar{\mathbf{F}}_{i})^2} \sqrt{\sum (\hat{\mathbf{D}}_{i}^{j} - \bar{\mathbf{D}}_{i})^2}},
\end{equation}
where $\bar{\mathbf{F}}_{i}$ and $\bar{\mathbf{D}}_{i}$ are the mean values of $\hat{\mathbf{F}}_{i}$ and $\hat{\mathbf{D}}_{i}$, respectively. $j$ is the position index of the matrices and $r \in \mathbb{R}^{C}$. 
As we know that $r$ is symmetric with respect to 0, where $r > 0$ indicates positive correlation and $r < 0$ signifies negative correlation. However, convolutional neural networks (CNNs) typically assign higher weight to features with high similarity and lower weight to those with less similarity instead of letting negative correlations cancel out positive ones. We thus apply a sigmoid function $\mathcal{F}_{\sigma}(\cdot)$ to compute the final correlation $\alpha$ of $\hat{\mathbf{F}}_i$ and $\hat{\mathbf{D}}_i$. Particularly, $\mathcal{F}_{\sigma}(\cdot)$ maps $r$ to a more compact and nonnegative interval of $[\frac{1}{1+e},\frac{1}{1+e^{-1}}]$. The fused result $\tilde{\mathbf{D}}_i$ is given by: 
\begin{subequations}
\begin{align}
    &\alpha=\mathcal{F}_{\sigma}(r), \quad
    \beta=1-\alpha,\\
    &\tilde{\mathbf{D}}_{i}=\alpha \hat{\mathbf{F}}_{i}+\beta \hat{\mathbf{D}}_{i}.
    \label{eq_fused_result}
\end{align}
\end{subequations}
Eq.~\eqref{eq_fused_result} adaptively selects regions of $\hat{\mathbf{F}}_{i}$ that exhibit a strong correlation with $\hat{\mathbf{D}}_{i}$ while preserving the original components of $\hat{\mathbf{D}}_{i}$ in areas with low correlation. This mechanism is especially effective for CNNs in probabilistic modeling, where negative correlation does not merely negate positive correlation but rather acts as a diminished weight in the computation. 
As a result, this approach effectively regulates the flow of information.  

Furthermore, the iterative process involving Eqs.~\eqref{eq_cf_input_dep}-\eqref{eq_fused_result} refines the depth representation. Specifically, $\tilde{\mathbf{D}}_{i}$ from Eq.~\eqref{eq_fused_result} is used to update $\hat{\mathbf{D}}_{i}$ in Eq.~\eqref{eq_cf_input_dep}, and this refinement is repeated for three iterations. Afterward, a $3 \times 3$ convolution $\mathcal{F}_{{\tau}_2}(\cdot)$ is applied to obtain $\mathbf{D}_{i+1}$, which serves as the input depth feature for the subsequent CF module:  
\begin{equation}  
\mathbf{D}_{i+1}=\mathcal{F}_{{\tau}_2}(\tilde{\mathbf{D}}_{i}).\label{eq_end}
\end{equation}  
Overall, by defining $\mathcal{F}_{\psi}(\cdot,\cdot)$ as the complete CF module, we can unify Eqs.~\ref{eq_begin}-\ref{eq_end} into the following expression: 
\begin{equation}  
    \mathbf{D}_{i+1}=\mathcal{F}_{\psi}(\mathbf{F}_{i},\mathbf{D}_{i}).\label{eq_overall_cf}  
\end{equation}  

Additionally, despite this fusion process, the depth feature often lacks geometric details while the prompt retains richer geometric structures. Moreover, misalignment issues persist even after resizing the prompt to match the resolution of the depth feature. To mitigate these challenges, we introduce a simple yet effective constraint designed to enhance geometric consistency: 
\begin{equation}
    \mathcal{L}_{cf} = \frac{1}{n} \| \mathcal{H}(\tilde{\mathbf{D}}_i) - \mathcal{H}(\hat{\mathbf{F}}_i) \|_2^2
    \label{eq_loss_cf},
\end{equation}
where $\mathcal{H}(\cdot)$ denotes a combination of a $1 \times 1$ convolution followed by normalization. It reduces the features to a single channel and normalizes them to the range $[0,1]$. $n$ is the number of valid pixels.

Finally, given the fused feature $\mathbf{D}_4 \in \mathbb{R}^{C \times H \times W}$ produced from the fourth CF module, we conduct a $3\times 3$ convolutional tail $\mathcal{F}_{{\tau}_3}(\cdot)$ to map the channel $C$ to 1, yielding the final depth prediction:
\begin{equation}
    \mathbf{Y}=\mathcal{F}_{{\tau}_3} (\mathbf{D}_{4}).
    \label{eq_output}  
\end{equation}

\begin{figure}[t]
 \centering
 \includegraphics[width=0.999\columnwidth]{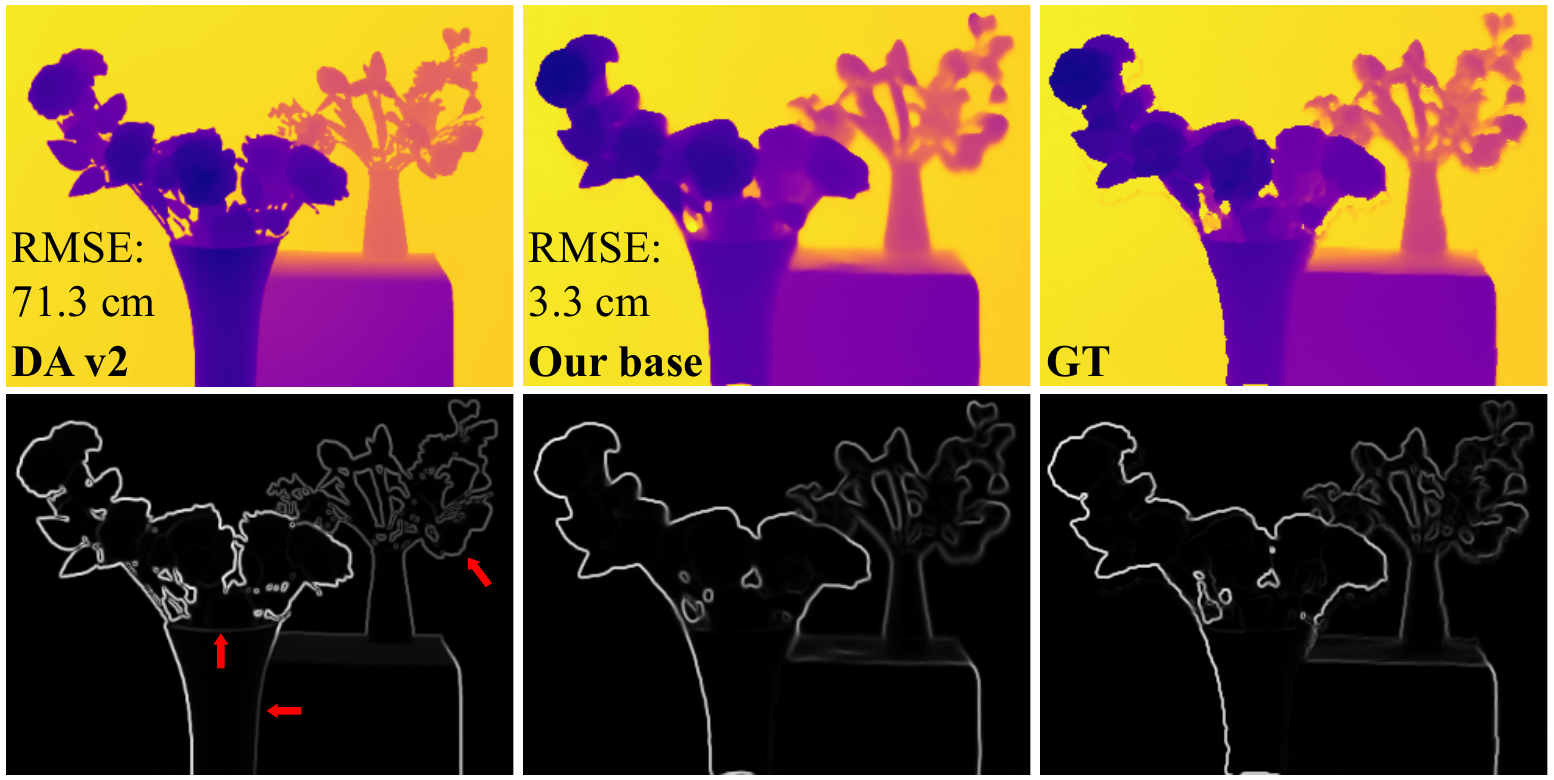}\\
  \vspace{-5pt}
 \caption{Visual comparisons of foundation models and traditional DSR methods. DA v2: Depth Anything v2 \cite{yang2025depth} (ViT-L encoder). 
 }\label{fig_gn}
  \vspace{0pt}
\end{figure}

\noindent \textbf{Gradient Regulation.}
As shown in Fig.~\ref{fig_pipeline}, the image branch utilizing a foundation model also produces a depth prediction denoted as $\mathbf{Y}'$. However, monocular depth estimation is inherently an ill-posed problem since models can only infer the relative depth instead of absolute measurements. Fortunately, as shown in Fig.~\ref{fig_gn}, the visual quality of $\mathbf{Y}'$ remains remarkably high despite its significant error. Notably, $\mathbf{Y}'$ exhibits exceptional clarity and sharp gradient, even surpassing the ground truth depth. 

In view of the large discrepancy between the depth values of $\mathbf{Y}'$ and the target $\mathbf{Y}$, directly fusing these results is not appropriate. 
Instead, we utilize the gradient representation of $\mathbf{Y}'$ as a constraint to regulate its edges. Specifically, we first adjust the depth discrepancy using Min-Max Normalization $\mathcal{N}(\cdot)$. We then calculate the first derivative to get their gradient representations: 
\begin{subequations}
\begin{align}
    &\mathcal{G}(\mathbf{Y})=\sqrt{{{\left( \frac{\partial \mathcal{N}(\mathbf{Y})}{\partial x} \right)}^{2}}+{{\left( \frac{\partial \mathcal{N}(\mathbf{Y})}{\partial y} \right)}^{2}}},\\
    &\mathcal{G}(\mathbf{Y}')=\sqrt{{{\left( \frac{\partial \mathcal{N}(\mathbf{Y}')}{\partial x} \right)}^{2}}+{{\left( \frac{\partial \mathcal{N}(\mathbf{Y}')}{\partial y} \right)}^{2}}}.
\end{align}
\end{subequations}
Subsequently, we propagate the edge priors from the depth foundation model by minimizing their normalized gradient distances through:
\begin{equation}
    \mathcal{L}_{gr} = \frac{1}{n} \| \mathcal{G}(\mathbf{Y}) - \mathcal{G}(\mathbf{Y}') \|_1.
    \label{eq_loss_gr}
\end{equation}
This constraint allows us to improve the sharpness and overall quality of our estimated depth maps.

\subsection{Lagrangian Duality Optimization}
\label{sec_lagrangian_duality}
Given the HR depth prediction $\mathbf{Y}$ and the ground truth $\mathbf{Z}$, we use the commonly adopted reconstruction loss \cite{he2021towards,yan2022learning,metzger2023guided} for optimization: 
\begin{equation}
    \mathcal{L}_{rec} = \frac{1}{n} \| \mathbf{Y} - \mathbf{Z} \|_1.
    \label{eq_loss_rec}
\end{equation}

By integrating Eq.~\eqref{eq_loss_rec} with the two prompts from Eq.~\eqref{eq_loss_cf} and Eq.~\eqref{eq_loss_gr}, we can further enhance the optimization of the target depth. Unlike previous DSR methods~\cite{tang2021bridgenet,sun2021learning,zhao2023spherical,wang2024sgnet} that introduce additional loss terms as auxiliary constraints, we adopt the Lagrangian duality theory~\cite{boyd2004convex,palomar2006tutorial} to translate the optimization problem into a constrained form. Specifically, based on Eqs.~\eqref{eq_loss_cf} and~\eqref{eq_loss_gr}, the \textit{primal problem} can be written as:
\begin{equation}
    \underset{\omega }{\mathop{\min }}\,\mathcal{L}_{rec}, \quad  \text{s.t.} \quad  \mathcal{L}_{cf}< {\varepsilon }_{1}, \ \mathcal{L}_{gr}< {\varepsilon }_{2},
    \label{eq_primal_problem}
\end{equation}
where $\omega $ indicates the network weight of our DuCos. ${\varepsilon }_{1}$ and ${\varepsilon }_{2}$ are small positive values approaching zero, signifying that errors have become negligible. Next, we introduce two Lagrangian multipliers $\lambda$ and $\mu$ to formulate the Lagrangian function 
as:
\begin{equation}
\begin{split}
    &\mathcal{L}(\omega, \lambda, \mu) 
    = \underbrace{\mathcal{L}_{rec}}_{\text{target}} + \underbrace{\lambda \mathcal{L}_{cf} + \mu \mathcal{L}_{gr}}_{\text{Lag. const. term}}\\
    &= \frac{1}{n} \| \mathbf{Y} - \mathbf{Z} \|_1 + \frac{\lambda}{n} \| \mathcal{H}(\tilde{\mathbf{D}}_i) - \mathcal{H}(\hat{\mathbf{F}}_i) \|_2^2 \\
    &+ \frac{\mu}{n} \| \mathcal{G}(\mathbf{Y}) - \mathcal{G}(\mathbf{Y}') \|_1 ,
    \label{eq_loss_lagrangian}
\end{split}
\end{equation}
where $\lambda \geq 0$ and $\mu \geq 0$ according to the Karush-Kuhn-Tucker (KKT) conditions~\cite{kuhn1951nonlinear}. We then minimize the Lagrangian function to get the dual function:
\begin{equation}
    \mathcal{D}(\lambda, \mu) =\underset{\omega }{\mathop{\min }}\, \mathcal{L}(\omega, \lambda, \mu).
    \label{eq_dual_function}
\end{equation}
Subsequently, the \textit{dual problem} corresponding to the primal problem in Eq.~\eqref{eq_primal_problem} is formulated by maximizing Eq.~\eqref{eq_dual_function}:
\begin{equation}
    \underset{\lambda, \mu}{\mathop{\max }}\,\mathcal{D}(\lambda, \mu)=\underset{\lambda, \mu}{\mathop{\max }}\,\underset{\omega }{\mathop{\min }}\,\mathcal{L}(\omega , \lambda , \mu). 
    \label{eq_dual_problem}
\end{equation}
The dual optimization in Eq.~\eqref{eq_dual_problem} consists of two key steps: 
\textbf{\textit{Step 1}.}
Fix $\lambda$ and $\mu$, minimize the Lagrangian function to obtain the optimal network weight:
\begin{equation}
    {\omega}^{*}(\lambda, \mu) = 
    \arg \underset{\omega}{\mathop{\min }}\,(\mathcal{L}_{rec} + \lambda \mathcal{L}_{cf} + \mu \mathcal{L}_{gr}).
\end{equation}
\textbf{\textit{Step 2}.}
Adjust $\lambda$ and $\mu$, maximize the dual function:
\begin{equation}
    {\lambda}^{*}, {\mu}^{*} = \arg \underset{\lambda, \mu}{\mathop{\max }}\,\mathcal{D}(\lambda, \mu).
\end{equation}
Alg.~\ref{alg_dps} gives the details of the dual problem optimization in Eq.~\eqref{eq_dual_problem}. The Lagrangian duality optimization allows for the flexible integration of multiple constraint conditions and reconstruction objectives. The optimization process is simplified by relaxing the constraints, providing an exact solution when the loss function is convex and an approximate solution when the loss function is non-convex.

\begin{algorithm}[t]
\DontPrintSemicolon
  \SetAlgoLined
  \small
      \KwIn {$\mathbf{Y}, \mathbf{Z}, \tilde{\mathbf{D}}_{i}, \hat{\mathbf{D}}_{i}, \mathbf{Y}'
      $}

        Initialize learning rate $\eta_\omega =1\text{e-}5$, $\lambda =0.01$, $\mu =0.05$, 
        step-length $\eta_\lambda =\eta_\mu=0.01$ \;
        
        Set maximum epoch $T$  \;
        
        \textbf{for} epoch $t=1$ to $T$ \textbf{do} \;
        
        \hspace{12pt} \textbf{Step 1: Fix $\lambda$ and $\mu$, optimize $\omega$} \;

        \hspace{12pt} Total Lag. loss: 
        $\mathcal{L}(\omega, \lambda, \mu) = \mathcal{L}_{rec} + \lambda \mathcal{L}_{cf} + \mu \mathcal{L}_{gr}$
        \;

        \hspace{12pt} Gradients of $\omega$: $\nabla_{\omega} \mathcal{L}(\omega, \lambda, \mu)$ \;

        \hspace{12pt} Update $\omega$: 
        $\omega \leftarrow \omega - \eta_\omega \nabla_{\omega} \mathcal{L}(\omega, \lambda, \mu)$ \;

        \hspace{12pt} \textbf{Step 2: Optimize $\lambda$ and $\mu$} \;

        \hspace{12pt} Update $\eta_\lambda$, $\eta_\mu$: $\eta_\lambda \leftarrow \eta_\lambda(1-\frac{t}{T})$, \ $\eta_\mu \leftarrow \eta_\mu(1-\frac{t}{T})$ \;

        \hspace{12pt} Update $\lambda$, $\mu$:
        $\lambda \leftarrow \lambda + \eta_\lambda \mathcal{L}_{cf}, \ \mu \leftarrow \mu + \eta_\mu \mathcal{L}_{gr} $
        \;

        \hspace{12pt} Ensure $\lambda $, $\mu $ to satisfy KKT conditions: 

        \hspace{12pt} $\lambda \leftarrow \text{clamp}(\lambda, \min=0)$, $\mu \leftarrow  \text{clamp}(\mu, \min=0)$ \;

    \textbf{end for}  \;
    
    \KwOut {Optimized parameters $\omega$, $\lambda$, $\mu$}
  \caption{Dual Problem Solving}
  \label{alg_dps}
\end{algorithm}

\begin{figure*}[t]
 \centering
 
 \includegraphics[width=1.87\columnwidth]{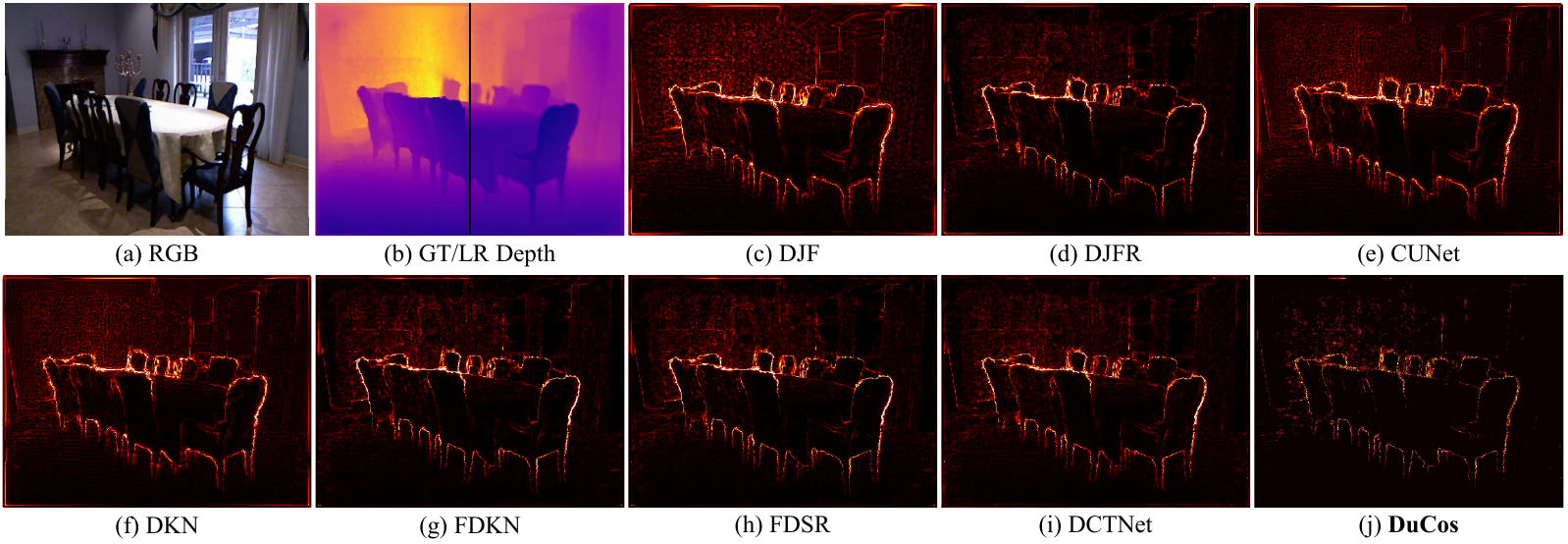}\\
 \vspace{-8pt}
 \caption{Error map comparisons on the synthetic NYU v2 dataset ($\times 4$), where warmer colors indicate higher errors.}\label{fig_vis_syn_nyu}
 \vspace{-4pt}
\end{figure*}

\begin{table*}[t]
\centering
\renewcommand\arraystretch{1.0}
\resizebox{0.894\linewidth}{!}{
\begin{tabular}{l|c|ccccccccccccccc}
\toprule 
\multirow{2}{*}{Method} &\multirow{2}{*}{Scale}   &\multicolumn{3}{c}{Middlebury}   &\multicolumn{3}{c}{Lu}  &\multicolumn{3}{c}{NYU v2}   &\multicolumn{3}{c}{RGB-D-D}  &\multicolumn{3}{c}{TOFDSR}\\ 
\cmidrule(lr){3-5}\cmidrule(lr){6-8}\cmidrule(lr){9-11}\cmidrule(lr){12-14} \cmidrule(lr){15-17} 
 & &RMSE &MAE &$\delta _{1}  $     &RMSE &MAE &$\delta _{1}  $  
 &RMSE &MAE &$\delta _{1}  $     &RMSE &MAE &$\delta _{1}  $ &RMSE &MAE &$\delta _{1}  $\\ \midrule
DJF~\cite{li2016deep}           &\multirow{9}{*}{$\times 2$}  &1.14 &0.63 &98.56       &1.08 &0.44 &99.00               &2.32 &0.83 &99.49      &1.18 &0.36 &99.64   &2.99 &0.66 &98.74   \\
DJFR~\cite{li2019joint}         &                             &1.22 &0.58 &98.42       &1.39 &0.45 &98.94               &1.87 &0.51 &99.68      &0.98 &0.27 &99.74   &1.58 &0.28 &99.57   \\
CUNet~\cite{deng2020deep}       &                             &1.01 &0.58 &98.78       &1.06 &0.50 &99.10               &\underline{1.50} &\underline{0.46} &\underline{99.81}      &\underline{0.84} &0.25 &\underline{99.81}   &1.69 &0.33 &99.49   \\
FDKN~\cite{kim2021deformable}   &                             &1.35 &0.62 &98.39       &1.59 &0.49 &98.91               &2.05 &0.50 &99.63      &1.04 &0.26 &99.70   &1.85 &0.36 &99.32   \\
DKN~\cite{kim2021deformable}    &                             &1.32 &0.62 &98.48       &1.48 &0.48 &99.00               &1.95 &0.49 &99.66      &0.98 &\underline{0.24} &99.74   &1.87 &0.36 &99.33   \\
FDSR~\cite{he2021towards}       &                             &\underline{0.94} &\underline{0.51} &98.89       &\underline{0.94} &\underline{0.39} &99.14               &1.74 &0.57 &99.70      &0.91 &0.26 &99.76   &1.55 &0.40 &99.24 \\
DCTNet~\cite{zhao2022discrete}  &                             &1.01 &0.53 &\underline{98.90}       &1.03 &0.59 &\underline{99.36}               &1.59 &0.59 &99.76      &0.89 &0.28 &99.79   &\underline{0.80} &\underline{0.23} &\underline{99.87}   \\
DADA~\cite{metzger2023guided}   &                             &1.28 &0.60 &98.47       &1.45 &0.48 &99.07               &2.00 &0.54 &99.65      &1.04 &0.27 &99.71   &1.73 &0.37 &99.26   \\
\textbf{DuCos}                  &              &\textbf{0.81} &\textbf{0.46} &\textbf{99.28}       &\textbf{0.65} &\textbf{0.26} &\textbf{99.68}           &\textbf{1.20} &\textbf{0.37} &\textbf{99.88}      &\textbf{0.74} &\textbf{0.22} &\textbf{99.85}   &\textbf{0.52} &\textbf{0.13} &\textbf{99.94} \\ 
\midrule
DJF~\cite{li2016deep}           &\multirow{9}{*}{$\times 4$} &1.93 &1.11 &97.11                               &1.93 &1.11 &98.20              &3.60 &1.67 &99.05                  &1.63 &0.74 &99.39       &3.75 &1.14 &98.18       \\
DJFR~\cite{li2019joint}         &                             &1.83 &1.04 &96.91                               &1.85 &1.02 &98.36              &3.27 &1.14 &99.04                  &1.54 &0.48 &99.35       &2.97 &0.58 &99.00       \\
CUNet~\cite{deng2020deep}       &                             &\underline{1.61} &0.98 &97.79                   &\underline{1.73} &0.97 &98.48              &3.22 &1.44 &99.21                  &1.52 &0.65 &99.43       &3.57 &1.13 &98.19       \\
FDKN~\cite{kim2021deformable}   &                             &1.80 &0.80 &97.69                               &2.11 &0.58 &98.53              &3.28 &0.93 &99.28                  &1.56 &0.42 &99.42       &2.80 &0.58 &98.96       \\
DKN~\cite{kim2021deformable}    &                             &1.77 &0.78 &97.77                               &2.05 &\underline{0.56} &\underline{98.58}  &3.15 &0.90 &99.34                  &1.50 &0.41 &99.49       &2.73 &0.56 &99.04    \\
FDSR~\cite{he2021towards}       &                             &1.72 &0.84 &97.69                               &1.95 &\underline{0.56} &98.52  &2.94 &\underline{0.85} &\underline{99.39}                  &\underline{1.41} &\underline{0.40} &\underline{99.51}       &\underline{2.41} &\underline{0.51} &\underline{99.07}    \\
DCTNet~\cite{zhao2022discrete}  &                             &1.66 &\underline{0.77} &\underline{97.86}       &1.85 &0.57 &98.56              &\underline{2.90} &0.99 &99.37                  &1.49 &0.44 &99.45       &2.86 &0.59 &98.98    \\
DADA~\cite{metzger2023guided}   &                             &1.82 &0.83 &97.45                               &2.12 &0.66 &98.35              &3.08 &0.96 &99.33                  &1.63 &0.46 &99.39       &2.85 &0.68 &98.57      \\
\textbf{DuCos}  &  &\textbf{1.45} &\textbf{0.68} &\textbf{98.30}  &\textbf{1.38}  &\textbf{0.41}  &\textbf{99.08}  &\textbf{2.60} &\textbf{0.81} &\textbf{99.49}  &\textbf{1.27} &\textbf{0.36} &\textbf{99.60}   &\textbf{2.09} &\textbf{0.33} &\textbf{99.58} \\ 
\midrule
DJF~\cite{li2016deep}           &\multirow{9}{*}{$\times 8$} &3.09 &1.46 &94.10                          &3.58 &1.22 &95.60            &5.56 &2.30 &97.70                  &2.58 &0.96 &98.29           &5.59 &1.71 &96.29  \\
DJFR~\cite{li2019joint}         &                             &2.82 &1.25 &95.31                          &3.24 &1.07 &95.86            &5.20 &1.94 &98.21                  &2.61 &0.93 &98.25           &5.11 &1.35 &97.16  \\
CUNet~\cite{deng2020deep}       &                             &2.86 &1.46 &94.44                          &2.85 &1.25 &96.32            &5.50 &2.23 &97.78                  &2.35 &0.92 &98.60           &5.14 &1.64 &96.67   \\
FDKN~\cite{kim2021deformable}   &  &2.51 &1.13 &\underline{96.26}   &\textbf{2.67} &\underline{0.85} &\underline{97.39}  &4.93 &\underline{1.67} &\underline{98.65}    &\underline{2.25} &\underline{0.70} &\underline{98.86}        &\underline{4.40} &0.96 &\underline{98.15}      \\
DKN~\cite{kim2021deformable}    &                             &2.43 &\underline{1.12} &\underline{96.26}              &2.88 &0.88 &97.13            &4.88 &1.71 &98.62                  &2.33 &0.74 &98.76        &4.54 &1.06 &97.87      \\
FDSR~\cite{he2021towards}       &                             &\underline{2.41} &1.13 &96.24  &\underline{2.69} &0.86 &97.35            &\underline{4.82} &1.69 &98.57    &\underline{2.25} &0.73 &98.72   &\textbf{4.28} &\underline{0.95} &98.06          \\
DCTNet~\cite{zhao2022discrete}  &                             &2.75 &1.31 &95.25                          &3.07 &1.18 &96.38            &4.90 &2.12 &98.28                  &2.47 &0.92 &98.60      &5.38 &1.57 &97.01       \\
DADA~\cite{metzger2023guided}   &                             &2.77 &1.27 &95.01                          &3.76 &1.17 &96.01            &4.83 &1.86 &98.34                  &2.81 &0.93 &98.29      &5.86 &1.64 &96.15     \\
\textbf{DuCos}  &  &\textbf{2.23} &\textbf{0.97} &\textbf{97.07}     &\textbf{2.67} &\textbf{0.72} &\textbf{97.86}    &\textbf{4.61} &\textbf{1.52} &\textbf{98.86}    &\textbf{2.23} &\textbf{0.66} &\textbf{98.89}  &4.60 &\textbf{0.92} &\textbf{98.30}\\ 

\midrule
DJF~\cite{li2016deep}           &\multirow{9}{*}{$\times16$} &5.50 &2.92 &84.66       &6.53 &2.69 &88.91            &9.82 &4.73 &93.05                  &4.46 &2.04 &94.99        &8.19 &3.49 &90.79      \\
DJFR~\cite{li2019joint}         &                             &5.16 &2.61 &86.39       &6.46 &2.38 &89.66            &9.50 &4.28 &93.63                  &4.36 &1.84 &95.48        &8.06 &3.06 &92.20      \\
CUNet~\cite{deng2020deep}       &                             &4.72 &2.40 &88.45       &5.63 &2.17 &91.33            &8.63 &3.88 &94.55                  &3.81 &1.58 &96.36        &7.36 &2.73 &93.40      \\
FDKN~\cite{kim2021deformable}   &                             &4.42 &2.10 &90.80       &5.48 &1.91 &92.24            &7.97 &3.43 &95.76                  &3.71 &1.45 &96.81        &7.16 &2.29 &94.78      \\
DKN~\cite{kim2021deformable}    &                             &4.17 &1.97 &91.67       &5.44 &1.90 &91.94            &7.70 &3.28 &96.06                  &\underline{3.70} &1.39 &97.06        &7.24 &2.20 &95.26      \\
FDSR~\cite{he2021towards}       &                             &\underline{3.97} &\underline{1.81} &\underline{92.52}       &\underline{5.23} &\underline{1.67} &\underline{93.44}            &\textbf{7.29} &\underline{2.91} &\underline{96.86}                  &\textbf{3.44} &\underline{1.24} &\underline{97.37}        &\textbf{6.85} &\underline{1.91} &\underline{95.96}     \\
DCTNet~\cite{zhao2022discrete}  &                             &5.07 &2.59 &86.91       &5.83 &2.25 &90.19            &9.10 &4.24 &94.04                  &4.13 &1.66 &96.30        &8.04 &2.77 &93.45     \\
DADA~\cite{metzger2023guided}   &                             &4.11 &2.06 &90.09       &6.19 &2.22 &90.95            &7.99 &3.54 &95.64                  &4.01 &1.59 &96.71        &7.79 &2.70 &93.25     \\
\textbf{DuCos}   &   &\textbf{3.96} &\textbf{1.69} &\textbf{93.52}    &\textbf{5.18} &\textbf{1.59} &\textbf{93.89}    &\underline{7.37} &\textbf{2.82} &\textbf{97.12}      &\textbf{3.44} &\textbf{1.16} &\textbf{97.61}   &\underline{6.90} &\textbf{1.77} &\textbf{96.51}\\ 
\bottomrule
\end{tabular}}
\vspace{-8pt}
\caption{Quantitative comparisons on the synthetic DSR benchmarks. The \textbf{best} and \underline{second-best} results are highlighted. 
}\label{tab_syn}
\vspace{-11pt}
\end{table*}

\section{Experiment}
\subsection{Dataset}
With the advancement of computer vision, training with simulated data has become a crucial approach to reducing data acquisition costs. Thus, we retrain all methods on the fully simulated Hypersim~\cite{roberts2021hypersim} dataset and evaluate them on the test sets of various DSR datasets: 
TOFDSR~\cite{yan2024tri}, RGB-D-D~\cite{he2021towards}, NYUv2~\cite{silberman2012indoor}, Middlebury~\cite{hirschmuller2007evaluation,scharstein2007learning}, and Lu~\cite{lu2014depth}. The high-quality RGB-D data of Hypersim are produced from 77,400 images across 461 indoor scenes, offering per-pixel geometric labels along with comprehensive scene geometry, material properties, and lighting information. For DSR training, we choose 2,000 RGB-D pairs. Building on previous approaches~\cite{zhao2022discrete,he2021towards,zhao2023spherical}, we use bicubic interpolation to generate LR depth from depth annotations across these five datasets. Additionally, TOFDSR and RGB-D-D also provide real-world LR depth data, enabling the evaluation of the generalization capabilities of the different models. Refer to our appendix for metrics and implementation details. 

\begin{figure*}[t]
\centering
\includegraphics[width=1.87\columnwidth]{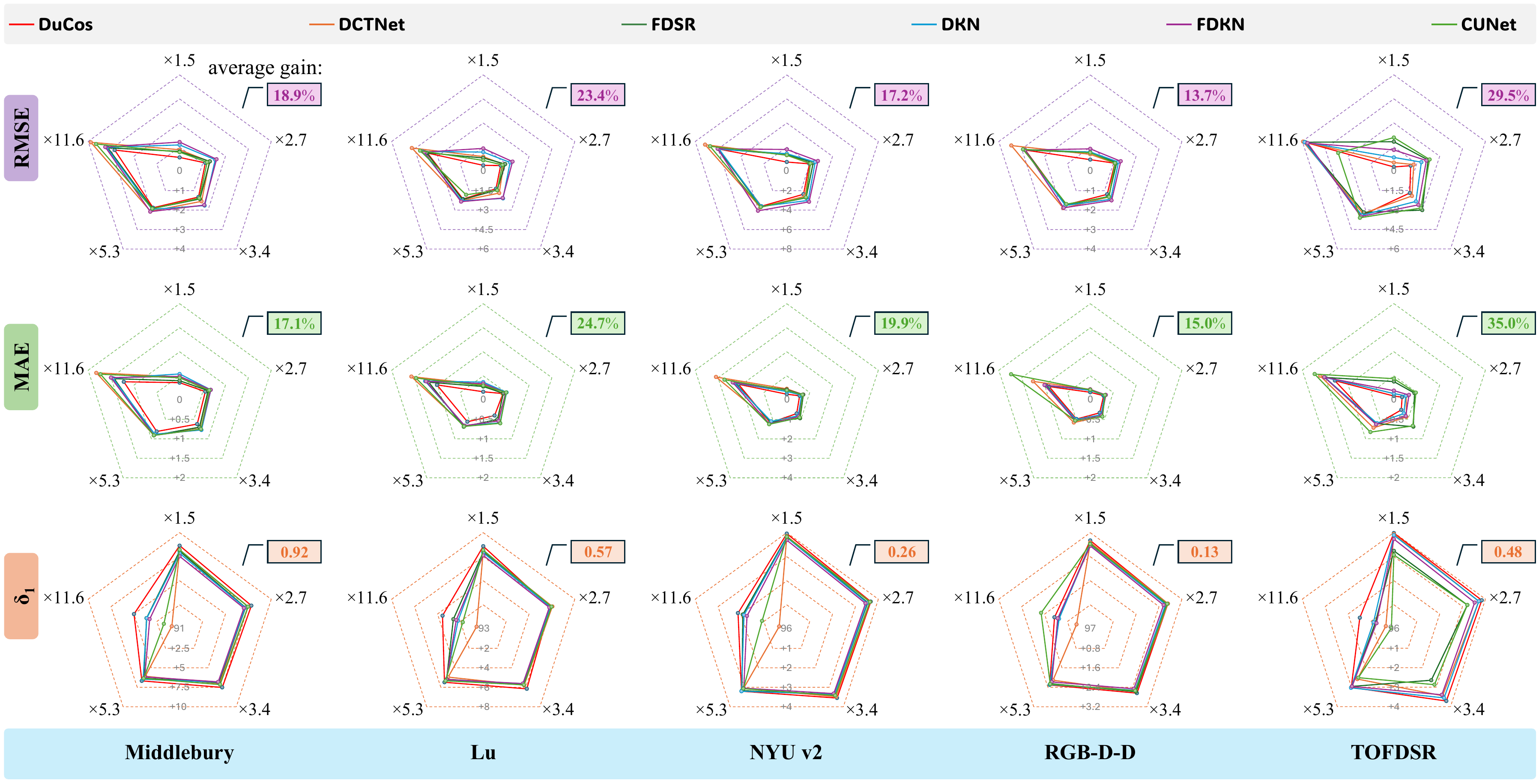}\\
\vspace{-9pt}
\caption{Comparisons with arbitrary scaling factors on the synthetic DSR datasets. Refer to our appendix for more details.}\label{fig_arbitrary}
\vspace{-6pt}
\end{figure*}

\subsection{Comparison with State-of-the-arts}
\textbf{Synthetic DSR.} 
We first evaluate our proposed DuCos on widely used synthetic DSR datasets. As shown in Tab.~\ref{tab_syn}, DuCos consistently delivers superior or highly competitive performance on various datasets and scales. Particularly, in the $\times 2$ case, our DuCos achieves the lowest RMSE and MAE while maintaining the highest $\delta _1$, outperforming second-best methods by an average of 29.1\% in RMSE and 29.6\% in MAE on all five datasets. Likewise, in the case of $\times 4$, our DuCos continues to demonstrate its superiority, exceeding the second-best approaches by 14.9\% in RMSE and 24.1\% in MAE, respectively. 
As the scale increases to $\times 8$ and $\times 16$, the performance of all methods naturally degrades due to the loss of information in depth maps with lower resolution. However, DuCos remains highly effective, preserving lower errors and higher accuracy. Fig.~\ref{fig_vis_syn_nyu} provides a visual comparison with the latest approaches, where the error maps highlight the ability of our DuCos to reconstruct depth with higher precision and thus it captures more accurate boundaries.

\begin{table*}[t]
\centering
\scriptsize
\renewcommand\arraystretch{1.0}
\resizebox{0.894\linewidth}{!}{
\begin{tabular}{l|ccccccccccccc}
\toprule 
\multirow{2}{*}{Method}   &\multicolumn{3}{c}{RGB-D-D}   &\multicolumn{3}{c}{TOFDSR}  &\multicolumn{3}{c}{RGB-D-D  w/ noise}   &\multicolumn{3}{c}{TOFDSR w/ noise} \\
\cmidrule(lr){2-4}\cmidrule(lr){5-7} \cmidrule(lr){8-10} \cmidrule(lr){11-13}
 &RMSE &MAE &$\delta _{1} $     &RMSE &MAE &$\delta _{1} $ 
 &RMSE &MAE &$\delta _{1} $     &RMSE &MAE &$\delta _{1} $\\ \midrule
DJF~\cite{li2016deep}                    &5.54 &3.43 &93.81                         &5.84 &2.13 &96.79                  &7.94 &5.16 &84.82              &11.45 &7.87 &67.95     \\  
DJFR~\cite{li2019joint}                  &5.52 &3.51 &93.58                         &5.72 &2.10 &97.03                  &7.50 &4.83 &86.25              &10.92 &7.39 &70.46     \\
CUNet~\cite{deng2020deep}                &5.84 &3.06 &94.75                         &6.04 &2.21 &96.46                  &6.69 &4.14 &89.36              &9.76 &5.86 &80.43           \\  
FDKN~\cite{kim2021deformable}            &5.37 &2.70 &96.05                         &5.77 &2.19 &97.33                  &6.66 &4.26 &90.09              &8.13 &4.66 &86.24           \\
DKN~\cite{kim2021deformable}             &5.08 &2.58 &96.28                         &5.50 &2.07 &97.54                  &6.50 &4.16 &90.04              &7.42 &4.29 &88.20           \\
FDSR~\cite{he2021towards}                &5.49 &3.10 &94.77                         &5.03 &1.67 &\underline{97.61}      &6.39  &4.07 &90.69             &6.31 &3.17 &92.74                       \\
DCTNet~\cite{zhao2022discrete}           &5.43 &3.29 &93.15                         &5.16 &2.10 &96.37                  &6.04  &\underline{3.79} &\underline{90.90}             &7.52 &4.50 &86.04           \\
SFG~\cite{yuan2023structure}             &\underline{3.88} &\underline{1.96} &\underline{97.09}         &\underline{4.52} &\underline{1.72} &97.45      &\underline{5.87} &\underline{3.79} &89.84                        &\underline{5.46} &\underline{2.89} &\underline{93.02}                        \\
\textbf{DuCos}                            &\textbf{3.68} &\textbf{1.54} &\textbf{97.87}          &\textbf{4.29} &\textbf{1.15} &\textbf{98.67}   &\textbf{4.14} &\textbf{2.00} &\textbf{96.76}           &\textbf{5.20} &\textbf{2.20} &\textbf{96.44} \\  
\bottomrule
\end{tabular}}
 \vspace{-8pt}
\caption{Quantitative comparisons on the real-world DSR datasets, including RGB-D-D, TOFDSR and their noisy patterns.}\label{tab_real}
 \vspace{-7pt}
\end{table*}

\begin{figure*}[!ht]
 \centering
 
 \includegraphics[width=1.87\columnwidth]{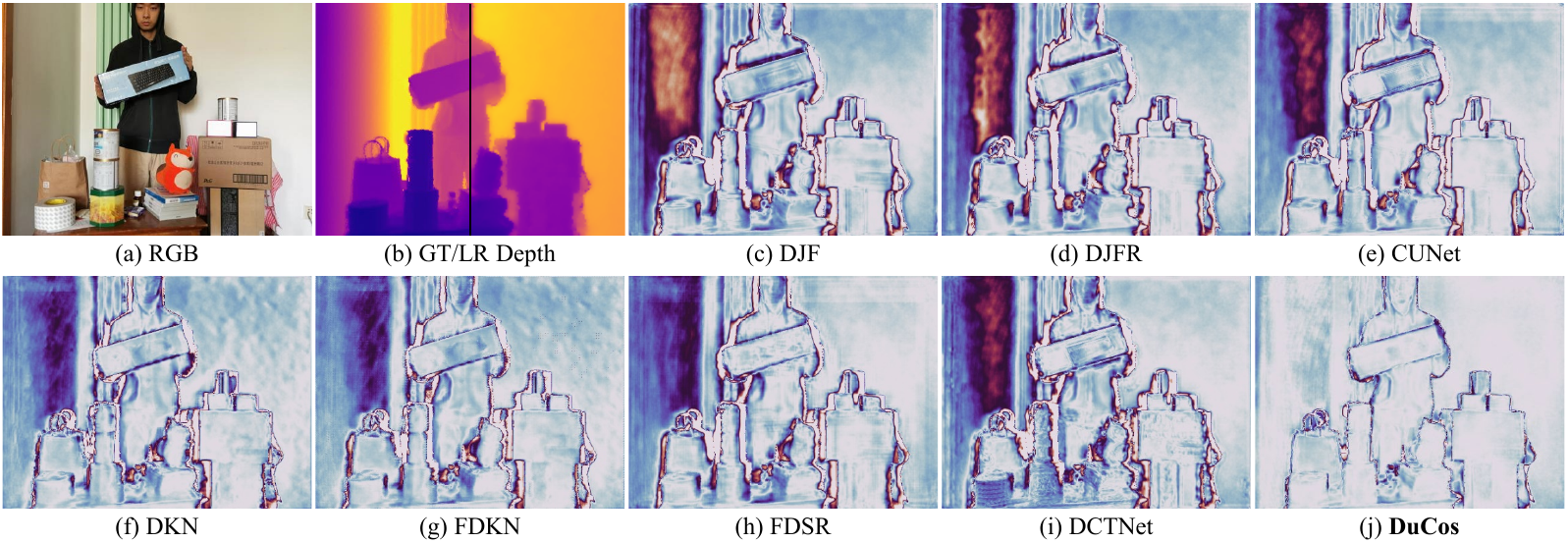}\\
  \vspace{-8pt}
 \caption{Error map comparisons on the real-world RGB-D-D dataset. Please see our appendix for more visualizations.}\label{fig_vis_real_rgbdd}
  \vspace{-8pt}
\end{figure*}

\noindent \textbf{Arbitrary-scale DSR.} 
In downstream applications, scaling factors are often variable or unknown. In addition to evaluating DSR methods at fixed integer scales, we thus also assess their performance on arbitrary scaling factors. As illustrated in Fig.~\ref{fig_arbitrary}, experiments are conducted at $\times 1.5$, $\times 2.7$, $\times 3.4$, $\times 5.3$, and $\times 11.6$ scales. Our DuCos consistently achieves outstanding results across all scales and datasets. Specifically, compared to all methods on these five scales, DuCos demonstrates significantly superior performance on RMSE, MAE, and $\delta_1$, with average improvements of 20.6\%, 22.3\%, and 0.47 percent points, respectively. These results further validate DuCos's effectiveness.

\noindent \textbf{Real-world DSR.} 
Since the introduction of FDSR \cite{he2021towards}, real-world DSR has gained increasing attention. Unlike integer or arbitrary scaling, the real-world setting reflects the practical challenges of DSR. Additionally, we assess model robustness by introducing noise to simulate external disturbances. As listed in Tab.~\ref{tab_real}, DuCos consistently outperforms all competing methods on both datasets. It surpasses the second-best SFG by 4.2 mm in MAE on RGB-D-D, and exceeds SFG by 5.7 mm on TOFDSR. Even under noisy conditions\footnote{Following SFG~\cite{yuan2023structure}, we applied Gaussian blur (standard deviation: 3.6) and Gaussian noise (mean: 0, standard deviation: 0.07) to LR depth.}, our DuCos maintains its advantage. On average, DuCos outperforms the suboptimal methods by 17.1\% in RMSE, 35.6\% in MAE, and 4.6 points in $\delta_1$, demonstrating strong robustness against noise. Fig.~\ref{fig_vis_real_rgbdd} shows that DuCos recovers a higher-quality depth map near the boundaries and textureless areas.

\noindent \textbf{Compressed DSR.} 
Due to the limitations of consumer-grade depth cameras and bandwidth constraints, restoring accurate depth from compressed sources is crucial. Following GDNet~\cite{zheng2024decoupling}, Tab.~\ref{tab_compressed} presents the $\times 8$ results. Our DuCos consistently outperforms all previous methods, achieving average improvements of 37.6\% in RMSE, 40.4\% in MAE, and 8.5 points in $\delta_1$ over the second-best methods. 
These significant improvements underscore the effectiveness and generalization of our approach in minimizing reconstruction error while substantially enhancing accuracy.

\begin{table}[t]
\centering
\huge
\renewcommand\arraystretch{1.0}
\resizebox{1.0\linewidth}{!}{
\begin{tabular}{l|ccccccccc}
\toprule 
\multirow{2}{*}{Method}   &\multicolumn{3}{c}{NYU v2}   &\multicolumn{3}{c}{RGB-D-D}  &\multicolumn{3}{c}{TOFDSR} \\
\cmidrule(lr){2-4}\cmidrule(lr){5-7}\cmidrule(lr){8-10}
&RMSE &MAE &$\delta _{1.05}$      &RMSE &MAE &$\delta _{1.05} $  
 &RMSE &MAE &$\delta _{1.05}$    \\ \midrule
DJF~\cite{li2016deep}             &119.41 &\underline{101.14} &\underline{8.53}      &\underline{34.48} &\underline{29.93} &16.26                                           &47.98 &44.30 &10.94     \\
DJFR~\cite{li2019joint}           &147.43 &121.73 &7.85                              &40.46 &34.49 &14.03                                           &50.14 &45.81 &8.59     \\
CUNet~\cite{deng2020deep}         &133.90 &108.96 &7.14                              &35.03 &30.67 &15.08                                           &50.25 &46.53 &8.96       \\
FDKN~\cite{kim2021deformable}     &187.53 &154.15 &4.26                              &42.37 &36.91 &14.04                                           &49.34 &44.88 &14.28       \\
DKN~\cite{kim2021deformable}      &186.10 &152.74 &4.33                              &40.38 &35.15 &14.40                                           &47.27 &43.34 &9.58       \\  
FDSR~\cite{he2021towards}         &118.76 &101.28 &8.39                  &35.88 &30.15 &\underline{20.64}       &\underline{45.96} &\underline{40.92} &\underline{20.26}       \\  
DCTNet~\cite{zhao2022discrete}    &\underline{118.45} &101.30 &8.20                  &37.85 &33.25 &12.82                                           &51.58 &48.41 &6.88       \\  
\textbf{DuCos}                     &\textbf{58.74}  &\textbf{46.00}  &\textbf{26.19}  &\textbf{24.64} &\textbf{21.13} &\textbf{26.45}                &\textbf{30.40} &\textbf{25.63} & \textbf{22.37}       \\ 
\bottomrule
\end{tabular}}
 \vspace{-8pt}
\caption{Quantitative comparisons of compressed DSR on NYU v2, RGB-D-D, and TOFDSR datasets. 
}\label{tab_compressed}
 \vspace{0pt}
\end{table}

\begin{figure}[t]
\centering
\includegraphics[width=0.999\columnwidth]{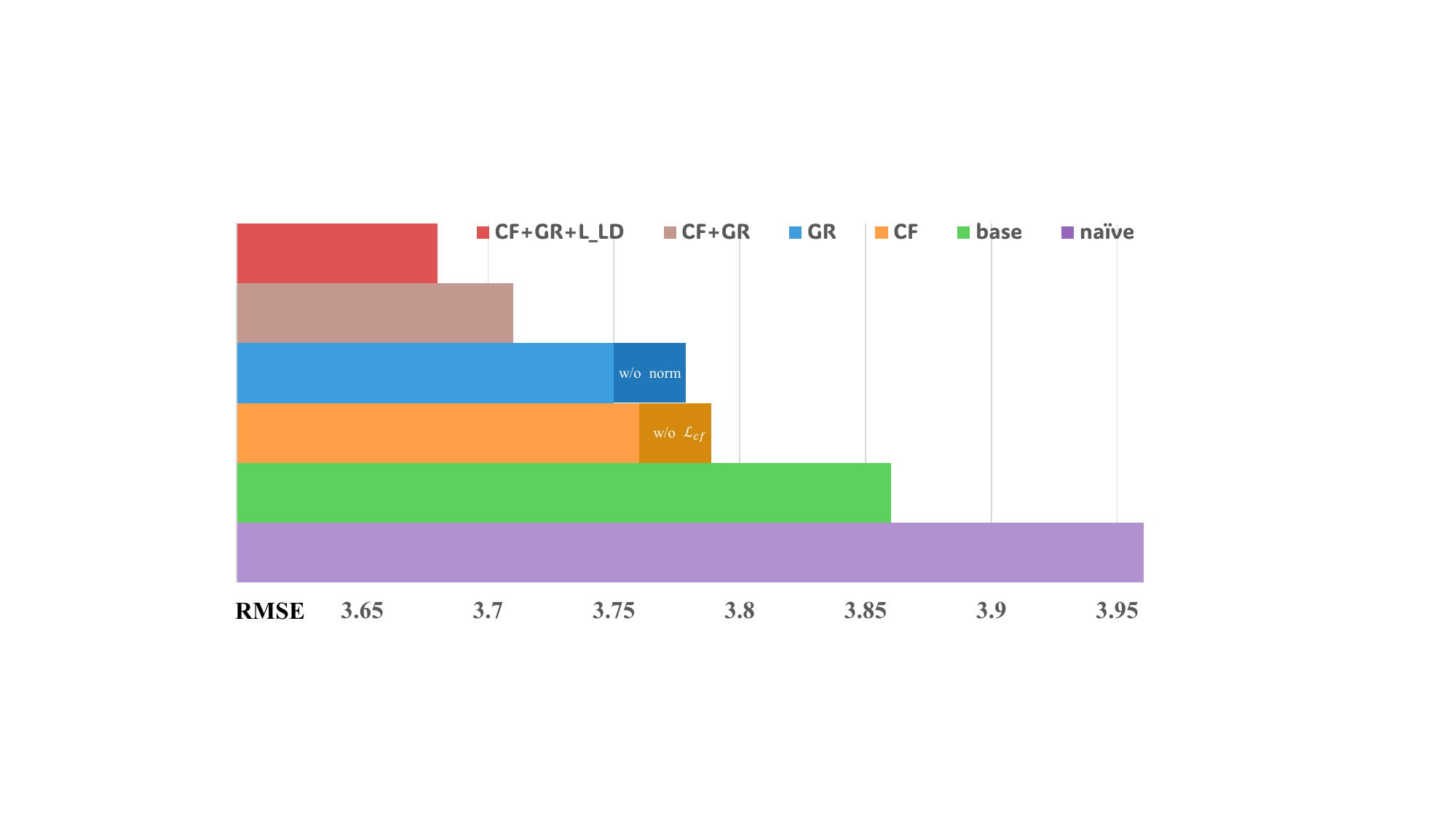}\\
\vspace{-7pt}
\caption{Ablation study of DuCos on the real-world RGB-D-D.}\label{fig_ablation}
\vspace{-6pt}
\end{figure}

\subsection{Ablation Study}
\textbf{DuCos designs.} 
Fig.~\ref{fig_ablation} presents the ablation results on the real-world RGB-D-D. The naïve model consists of separate image and depth branches, each using four residual groups \cite{zhang2018image} for feature extraction and fused by element-wise addition. Our introduction of depth foundation models as prompts reduces the RMSE from 3.97 cm to 3.86 cm. Based on this, CF improves performance by 0.1 cm, while GR yields a 0.11 cm improvement. Furthermore, combining CF and GR reduces the error of the base model by 0.15 cm. 
Finally, incorporating the LD strategy (red bar) yields the best performance, achieving an RMSE of 3.68 cm. 
Notably, the alignment loss $\mathcal{L}_{cf}$ in CF also leads to performance gains, reducing the RMSE from 3.82 cm to 3.76 cm. Additionally, normalization in GR mitigates discrepancies between the target depth and the relative depth from the foundation models, further reducing the error by 0.05 cm. In summary, each component contributes positively to the entire model.

\begin{table}[t]
\centering
\huge
\renewcommand\arraystretch{1.0}
\resizebox{1.0\linewidth}{!}{
\begin{tabular}{l|cc|cccccc}
\toprule 
\multirow{2}{*}{Prompt in DuCos}  & Params.  & Time  &\multicolumn{3}{c}{NYU v2}   &\multicolumn{3}{c}{Middlebury}  \\
\cmidrule(lr){2-3}\cmidrule(lr){4-6}\cmidrule(lr){7-9}
&(M)  &(ms) &RMSE &MAE &$\delta _{1.05}$ &RMSE &MAE &$\delta _{1.05}$      \\ \midrule
DepthFormer \cite{li2023depthformer} & 282.76  & 58.47  & \underline{7.27}  & 2.82  & 97.12  & 4.01  & 1.74  & 93.21  \\ 
UniDepth v2-S \cite{piccinelli2024unidepth} & \underline{87.21}  & 91.27  & 8.69  & 3.30  & 96.12 
 & 4.21  & 1.79  & 93.12  \\ 
UniDepth v2-L \cite{piccinelli2024unidepth} & 372.97  & 159.28  & 8.86  & 3.45  & 95.85  & 5.40 
 & 2.37  & 89.83  \\ 
DA v2-S \cite{yang2025depth} & \textbf{34.38} 
 & \textbf{23.53}  &  7.37  & 2.82  & 97.12 & 3.96  & \underline{1.69}  & 93.52  \\ 
DA v2-B \cite{yang2025depth} & 107.19  & \underline{31.24}  & 7.34  & \underline{2.79}  & \underline{97.14}  & \underline{3.94}  & \underline{1.69}  & \underline{93.61}  \\ 
DA v2-L \cite{yang2025depth} & 345.06  & 71.81  & \textbf{7.15}  & \textbf{2.74}  & \textbf{97.29}  & \textbf{3.64}  &  \textbf{1.55}  & \textbf{94.35}  \\ 
\bottomrule
\end{tabular}}
 \vspace{-8pt}
\caption{Ablation study of DuCos with different foundation models on the synthetic NYU v2 ($\times 16$) and Middlebury ($\times 16$).}\label{tab_ablation_fm}
 \vspace{0pt}
\end{table}

\begin{figure}[t]
\centering
\includegraphics[width=0.999\columnwidth]{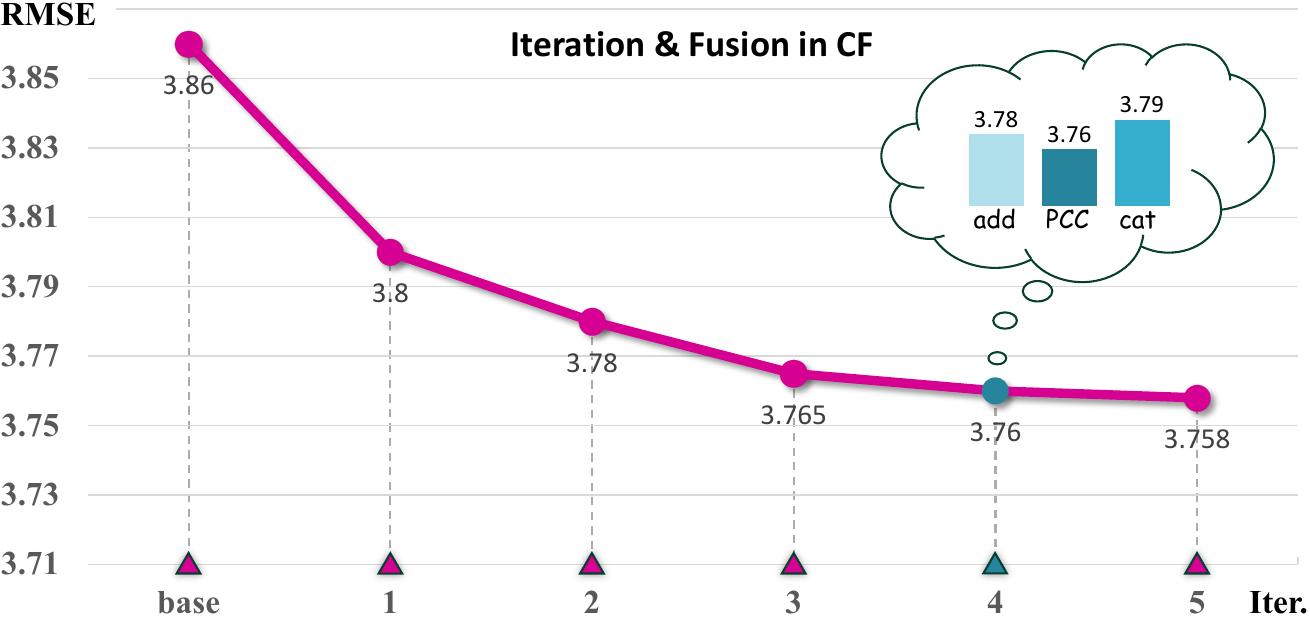}\\
\vspace{-6pt}
\caption{Ablation study of CF on the real-world RGB-D-D.}\label{fig_ablation_cf}
\vspace{-4pt}
\end{figure}

\noindent \textbf{Iteration \& fusion in CF.} 
Fig.~\ref{fig_ablation_cf} shows the ablation study of the CF module. As the number of iterations increases, the error progressively decreases. To balance performance and computational complexity, we set the default iteration count to 4. Furthermore, our findings indicate that the fusion based on the Pearson correlation coefficient (PCC) outperforms both addition and concatenation, achieving improvements of 0.02 cm and 0.03 cm, respectively.

\noindent \textbf{Depth Foundation Models.} 
Tab.~\ref{tab_ablation_fm} lists the ablation study of DuCos using different foundation models as prompts, including DepthFormer~\cite{li2023depthformer}, UniDepth v2~\cite{piccinelli2024unidepth}, and DA v2~\cite{yang2025depth}. The inference time is measured on a single 4090 GPU. The results indicate that heavier prompts generally yield better performance than their smaller counterparts. Among these models, DA v2-L achieves the highest performance. Considering the complexity-performance trade-off, we adopt DA v2-S as the default prompt for DuCos.

\section{Conclusion}
In this paper, we introduce DuCos, a novel DSR paradigm based on Lagrangian duality theory. By formulating DSR as a constrained optimization problem, our DuCos can restore global reconstruction with local geometric consistency, leading to better depth predictions. Furthermore, we pioneer the use of foundation models as prompts, leveraging correlative fusion and gradient regulation to enhance generalization across diverse scenarios. These prompts are seamlessly incorporated into the Lagrangian constraint term, creating a synergistic framework that combines theoretical rigor with practical adaptability. Extensive experiments demonstrate the superiority of our DuCos in terms of accuracy, robustness, and generalization. 


\section*{Acknowledgment}
This research/project is supported by the National Research Foundation, Singapore, under its NRF-Investigatorship Programme (Award ID. NRF-NRFI09-0008).

{
    \small
    \bibliographystyle{ieeenat_fullname}
    \bibliography{main}
}



\end{document}